\documentclass[a4paper,11pt]{article}
\usepackage[utf8]{inputenc}
\usepackage[T1]{fontenc} 
\usepackage{float} 
\usepackage{multirow}
\usepackage{cite}

\makeatletter
\gdef\@fpheader{ }
\gdef\@journal{ }
\makeatother
\RequirePackage{amsmath}
\RequirePackage{amssymb}
\RequirePackage{epsfig}
\RequirePackage{graphicx}
\RequirePackage[numbers,sort&compress]{natbib}
\RequirePackage{color}
\RequirePackage[colorlinks=true
,urlcolor=blue
,anchorcolor=blue
,citecolor=blue
,filecolor=blue
,linkcolor=blue
,menucolor=blue
,pagecolor=blue
,linktocpage=true
,pdfproducer=medialab
,pdfa=true
]{hyperref}

\newif\ifnotoc\notocfalse
\newif\ifemailadd\emailaddfalse
\newif\iftoccontinuous\toccontinuousfalse
\makeatletter
\def\@subheader{\@empty}
\def\@keywords{\@empty}
\def\@abstract{\@empty}
\def\@xtum{\@empty}
\def\@dedicated{\@empty}
\def\@arxivnumber{\@empty}
\def\@collaboration{\@empty}
\def\@collaborationImg{\@empty}
\def\@proceeding{\@empty}
\def\@preprint{\@empty}

\newcommand{\subheader}[1]{\gdef\@subheader{#1}}
\newcommand{\keywords}[1]{\if!\@keywords!\gdef\@keywords{#1}\else%
\PackageWarningNoLine{\jname}{Keywords already defined.\MessageBreak Ignoring last definition.}\fi}
\renewcommand{\abstract}[1]{\gdef\@abstract{#1}}
\newcommand{\dedicated}[1]{\gdef\@dedicated{#1}}
\newcommand{\arxivnumber}[1]{\gdef\@arxivnumber{#1}}
\newcommand{\proceeding}[1]{\gdef\@proceeding{#1}}
\newcommand{\xtumfont}[1]{\textsc{#1}}
\newcommand{\correctionref}[3]{\gdef\@xtum{\xtumfont{#1} \href{#2}{#3}}}
\newcommand\jname{JHEP}

\newcommand\preprint[1]{\gdef\@preprint{\hfill #1}}

\makeatother
\newcommand\note[2][]{%
\if!#1!%
\stepcounter{footnote}\footnotetext{#2}%
\else%
{\renewcommand\thefootnote{#1}%
\footnotetext{#2}}%
\fi}

\makeatletter
\newtoks\auth@toks
\renewcommand{\author}[2][]{%
  \if!#1!%
    \auth@toks=\expandafter{\the\auth@toks#2\ }%
  \else
    \auth@toks=\expandafter{\the\auth@toks#2$^{#1}$\ }%
  \fi
}
\makeatother
\makeatletter
\newtoks\affil@toks\newif\ifaffil\affilfalse
\newcommand{\affiliation}[2][]{%
\affiltrue
  \if!#1!%
    \affil@toks=\expandafter{\the\affil@toks{\item[]#2}}%
  \else
    \affil@toks=\expandafter{\the\affil@toks{\item[$^{#1}$]#2}}%
  \fi
}
\makeatother

\makeatletter
\newtoks\email@toks\newcounter{email@counter}%
\newcommand{\emailAdd}[1]{%
\emailaddtrue%
\ifnum\theemail@counter>0\email@toks=\expandafter{\the\email@toks, \@email{#1}}%
\else\email@toks=\expandafter{\the\email@toks\@email{#1}}%
\fi\stepcounter{email@counter}}
\newcommand{\@email}[1]{\href{mailto:#1}{\tt #1}}
\makeatother

\makeatletter
\newcommand*\collaboration[1]{\gdef\@collaboration{#1}}
\newcommand*\collaborationImg[2][]{\gdef\@collaborationImg{#2}}
\makeatletter

\newcommand\afterLogoSpace{\smallskip}
\newcommand\afterSubheaderSpace{\vskip3pt plus 2pt minus 1pt}
\newcommand\afterProceedingsSpace{\vskip21pt plus0.4fil minus15pt}
\newcommand\afterTitleSpace{\vskip23pt plus0.06fil minus13pt}
\newcommand\afterRuleSpace{\vskip23pt plus0.06fil minus13pt}
\newcommand\afterCollaborationSpace{\vskip3pt plus 2pt minus 1pt}
\newcommand\afterCollaborationImgSpace{\vskip3pt plus 2pt minus 1pt}
\newcommand\afterAuthorSpace{\vskip5pt plus4pt minus4pt}
\newcommand\afterAffiliationSpace{\vskip3pt plus3pt}
\newcommand\afterEmailSpace{\vskip16pt plus9pt minus10pt\filbreak}
\newcommand\afterXtumSpace{\par\bigskip}
\newcommand\afterAbstractSpace{\vskip16pt plus9pt minus13pt}
\newcommand\afterKeywordsSpace{\vskip16pt plus9pt minus13pt}
\newcommand\afterArxivSpace{\vskip3pt plus0.01fil minus10pt}
\newcommand\afterDedicatedSpace{\vskip0pt plus0.01fil}
\newcommand\afterTocSpace{\bigskip\medskip}
\newcommand\afterTocRuleSpace{\bigskip\bigskip}

\newlength{\affiliationsSep}
\newcommand\beforetochook{\pagestyle{myplain}\pagenumbering{roman}}

\DeclareFixedFont\trfont{OT1}{phv}{b}{sc}{11}

\renewcommand\maketitle{
\pagestyle{empty}
\thispagestyle{titlepage}
\setcounter{page}{0}
\noindent{\small\scshape\@fpheader}\@preprint\par

\afterLogoSpace
\if!\@subheader!\else\noindent{\trfont{\@subheader}}\fi
\afterSubheaderSpace
\if!\@proceeding!\else\noindent{\sc\@proceeding}\fi
\afterProceedingsSpace
{\LARGE\flushleft\sffamily\bfseries\@title\par}
\afterTitleSpace
\hrule height 1.5\p@%
\afterRuleSpace
\if!\@collaboration!\else
{\Large\bfseries\sffamily\raggedright\@collaboration}\par
\afterCollaborationSpace
\fi
\if!\@collaborationImg!\else
{\normalsize\bfseries\sffamily\raggedright\@collaborationImg}\par
\afterCollaborationImgSpace
\fi
{\bfseries\raggedright\sffamily\the\auth@toks\par}
\afterAuthorSpace
\ifaffil\begin{list}{}{%
\setlength{\leftmargin}{0.28cm}%
\setlength{\labelsep}{0pt}%
\setlength{\itemsep}{\affiliationsSep}%
\setlength{\topsep}{-\parskip}}
\itshape\small%
\the\affil@toks
\end{list}\fi
\afterAffiliationSpace
\ifemailadd 
\noindent\hspace{0.28cm}\begin{minipage}[l]{.9\textwidth}
\begin{flushleft}
\textit{E-mail:} \the\email@toks
\end{flushleft}
\end{minipage}
\else 
\PackageWarningNoLine{\jname}{E-mails are missing.\MessageBreak Plese use \protect\emailAdd\space macro to provide e-mails.}
\fi
\afterEmailSpace
\if!\@xtum!\else\noindent{\@xtum}\afterXtumSpace\fi
\if!\@abstract!\else\noindent{\renewcommand\baselinestretch{.9}\textsc{Abstract:}}\ \@abstract\afterAbstractSpace\fi
\if!\@keywords!\else\noindent{\textsc{Keywords:}} \@keywords\afterKeywordsSpace\fi
\if!\@arxivnumber!\else\noindent{\textsc{ArXiv ePrint:}} \href{http://arxiv.org/abs/\@arxivnumber}{\@arxivnumber}\afterArxivSpace\fi
\if!\@dedicated!\else\vbox{\small\it\raggedleft\@dedicated}\afterDedicatedSpace\fi
\ifnotoc\else
\iftoccontinuous\else\newpage\fi
\beforetochook\hrule
\tableofcontents
\afterTocSpace
\hrule
\afterTocRuleSpace
\fi
\setcounter{footnote}{0}
\pagestyle{myplain}\pagenumbering{arabic}
} 

\renewcommand{\baselinestretch}{1.1}\normalsize
\divide\@tempdima\baselineskip
\@tempcnta=\@tempdima
\skip\@mpfootins = \skip\footins
\renewcommand{\@dotsep}{10000}

\newcommand\ps@myplain{
\pagenumbering{arabic}
\renewcommand\@oddfoot{\hfill-- \thepage\ --\hfill}
\renewcommand\@oddhead{}}
\let\ps@plain=\ps@myplain

\newcommand\ps@titlepage{\renewcommand\@oddfoot{}\renewcommand\@oddhead{}}

\numberwithin{equation}{section}

\renewcommand\section{\@startsection{section}{1}{\z@}%
                                   {-3.5ex \@plus -1.3ex \@minus -.7ex}%
                                   {2.3ex \@plus.4ex \@minus .4ex}%
                                   {\normalfont\large\bfseries}}
\renewcommand\subsection{\@startsection{subsection}{2}{\z@}%
                                   {-2.3ex\@plus -1ex \@minus -.5ex}%
                                   {1.2ex \@plus .3ex \@minus .3ex}%
                                   {\normalfont\normalsize\bfseries}}
\renewcommand\subsubsection{\@startsection{subsubsection}{3}{\z@}%
                                   {-2.3ex\@plus -1ex \@minus -.5ex}%
                                   {1ex \@plus .2ex \@minus .2ex}%
                                   {\normalfont\normalsize\bfseries}}
\renewcommand\paragraph{\@startsection{paragraph}{4}{\z@}%
                                   {1.75ex \@plus1ex \@minus.2ex}%
                                   {-1em}%
                                   {\normalfont\normalsize\bfseries}}
\renewcommand\subparagraph{\@startsection{subparagraph}{5}{\parindent}%
                                   {1.75ex \@plus1ex \@minus .2ex}%
                                   {-1em}%
                                   {\normalfont\normalsize\bfseries}}

\def\fnum@figure{\textbf{\figurename\nobreakspace\thefigure}}
\def\fnum@table{\textbf{\tablename\nobreakspace\thetable}}

\long\def\@makecaption#1#2{%
  \vskip\abovecaptionskip
  \sbox\@tempboxa{\small #1. #2}%
  \ifdim \wd\@tempboxa >\hsize
    \small #1. #2\par
  \else
    \global \@minipagefalse
    \hb@xt@\hsize{\hfil\box\@tempboxa\hfil}%
  \fi
  \vskip\belowcaptionskip}

\renewenvironment{thebibliography}[1]{%
\begin{oldthebibliography}{#1}%
\small%
\raggedright%
\setlength{\itemsep}{5pt plus 0.2ex minus 0.05ex}%
}%
{%
\end{oldthebibliography}%
}

\title{\boldmath Early Abnormal Detection of Sewage Pipe Network: Bagging of Various Abnormal Detection Algorithms}
\author[a]{Zhen-Yu Zhang,}
\author[a]{Guo-Xiang Shao,}
\author[a]{Chun-Ming Qiu,}
\author[a]{Yue-Jie Hou,}
\author[a]{En-Ming Zhao,}
\author[a,1]{and Chi-Chun Zhou,}\note{zhouchichun@dali.edu.cn}

\affiliation[a]{School of Engineering, Dali University, Dali, Yunnan 671003, PR China}
\date{May 2022}

\begin{document}
\abstract{Abnormalities of the sewage pipe network 
will affect the normal operation of the whole city. 
Therefore, it is important to detect the abnormalities early.
This paper propose an early abnormal-detection method. 
The abnormalities are detected by using the 
conventional algorithms, such as isolation forest algorithm,
two innovations are given:
(1) The current and historical data measured 
by the sensors placed in the sewage pipe network 
(such as ultrasonic Doppler flowmeter) are taken as the overall dataset, 
and then the general dataset is detected by using 
the conventional anomaly detection method to diagnose the anomaly of the data. 
The anomaly refers to the sample 
different from the others samples in the whole dataset.
Because the definition of anomaly is not through the algorithm, 
but the whole dataset, 
the construction of the whole dataset 
is the key to propose the early abnormal-detection algorithms. 
(2) A bagging strategy for a variety of 
conventional anomaly detection algorithms 
is proposed to achieve the early detection of anomalies 
with the high precision and recall. 
The results show that this method can achieve the early anomaly detection 
with the highest precision of 98.21\%, the recall rate 63.58\% 
and F1-score of 0.774.}

\maketitle
\flushbottom

\section{Induction}
In the city, sewage pipe network plays an important role.
For example, it treats the municipal wastewater ensuring that the city 
and groundwater are not polluted.
However, if the sewage pipe network is abnormal, 
it will affect the normal operation of the whole city 
and cause a series of problems. 
For example, the ignoring of abnormalities of sewage pipe network will 
lead to the explosion of the pipe and environmental pollution 
during a heavy rain or pipe aging.
Therefore, the early detection of the abnormalities is important,
because abnormalities can be checked in time 
to avoid problems caused by abnormalities.

Conventionally, the hardware based methods are used to detect 
abnormalities of sewage pipe network and they can be 
categoried into two types: 
(1) Detection of the data. 
This method is to detect the pipe network data 
through the flowmeter, 
analyze it and diagnose the abnormality.
For example: F larrarte \cite{2006Velocity} introduces the detection application 
and measurement strategy of flowmeter in sewage pipe network in detail. 
(2) The detection of sewage pipe network. 
This method is to detect and diagnose abnormalities 
by closed-circuit television detection (CCTV) and sonar. 
For example: Cezary et al. \cite{Cezary1998Inspection} Introduced the application of CCTV in sewage pipeline, 
discussed the operational process of pipeline inspection and improved it. 
Olga Duran et al. \cite{2002State} Proposed the defects of CCTV in pipe network detection, 
and proposed sensors based on infrared and microwave to detect sewers. 
Although the hardware based methods can efficiency detect, 
the methods have two main shortcomings: 
(1) Only based on hardware equipment, 
the timeliness of detection is not high, 
and it is difficult to feed back 
the abnormalities detected in the sewage pipe network timely. 
(2) The coverage of hardware equipment is limited, 
and it's not ensure that the hardware can 
detect all abnormalities in the sewage pipe network.

Therefore, in view of the shortcomings of hardware detection, 
at this stage, many scholars have established different models 
for the situation of sewage pipe network based on statistical 
or machine learning methods, 
and then combined with the data obtained by hardware detection 
to solve the problems of sewage pipe network. For example: 
micevski et al. \cite{2002Markov} constructed the structural degradation model 
of rainwater pipe network based on Markov chain and calibrated it with Bayes. 
Mashford et al. \cite{2011Prediction} established a pipe network 
condition prediction model based on support vector machine 
and applied it to the sewage pipe network in Adelaide, Australia. 
Harvey R R et al. \cite{2014Predicting} used CCTV detection data 
and random forest algorithm to predict 
the structural state of sewer network in Guelph, Canada. 
Tran et al. \cite{TRAN20071144,D2009Comparison}used neural network model 
to predict the structural and hydraulic conditions 
of sewage pipe network. 

These methods can effectively solve the shortcomings 
of hardware detection and improve 
the ability of detecting the sewage pipe network, 
but these methods need to model the sewage pipe network in advance. 
When the status of sewage pipe network is not clear, 
these methods can not be used for the detection of sewage pipe network.
Therefore, this paper proposes an efficient and low-cost unsupervised 
early abnormal-detection method, 
which can effectively detect the sewage pipe network without modeling. 
Although the algorithm use the existing 
conventional anomaly detection algorithm, 
there are two innovations: 
(1) The current and historical data measured 
by the sensors placed in the sewage pipe network 
(such as ultrasonic Doppler flowmeter) are taken as the overall dataset, 
and then the general dataset is detected by using 
the conventional anomaly detection method to diagnose the anomaly of the data. 
The anomaly refers to the sample 
different from the others samples in the whole dataset. 
Because the definition of anomaly is not through the algorithm, 
but the whole dataset, 
the construction of the whole dataset 
is not only the key to propose the early abnormal-detection algorithms, 
but also one of the important innovations of this paper. 
(2) In this paper, the conventional anomaly detection methods 
such as Isolation Forest, One Class SVM and Local Outlier Factor 
are used to detect the whole dataset. 
However, these traditional anomaly detection algorithms 
sometimes get unsatisfactory results. 
Therefore, this paper proposes bagging strategies of 
various anomaly detection algorithms to achieve 
high precision and recall rate data anomaly detection. 
In the experiment, the data are collected 
from the ultrasonic Doppler flowmeter installed 
at 136 points in the sewage pipe network of Erhai Lake, 
and then use these data to train the algorithm 
obtained after bagging strategy. 
The results show that this method can achieve early anomaly detection 
with the highest accuracy of $98.21\%$, the recall rate $63.58\%$ 
and F1-score of $0.774$.

This paper is organized as following:  
In Sec. 2, introduce the data collected from 
the 136 ultrasonic Doppler flowmeters 
in the sewage pipe network of Erhai Lake. 
And discuss the types of abnormalities of sewage pipe network. 
In Sec. 3, the construction of the overall data is introduced. 
Then, introduces the specific steps 
of anomaly detection based on current and historical data. 
In Sec. 4, Propose the method of this paper. 
The common anomaly detection algorithms are reviewed, 
and the strategy of bagging of various 
abnormal detection algorithms
is proposed.
In Sec. 5, the experiments are conducted and the results are analyzed.
The conclusions and discussions are given in Sec. 6.

\section{Types of abnormalities}
The intelligent monitoring project of The Pollution Control of Erhai Lake was established 
by the government of Dali City, China, 
in order to rectify the water pollution of Erhai Lake \cite{2016Response}. 
A total of 136 mobile monitoring points have been established around Erhai Lake, 
and ultrasonic Doppler flowmeter is 
used to detect the flow at the monitoring points to obtain the flow data. 
This paper studies these flow data,  
and summarizes five types of anomalies.

1) Mixing of external water sources

External water sources generally refer to rainwater, 
groundwater, etc. Taking the mixed rainwater as an example, 
when it rains, the rainwater flows into the sewage pipe network, 
and the sewage flow and liquid level rise sharply. 
Mixing with external water sources increases the load of sewage pipe network, 
which is easy to lead to damage of sewage pipe network and environmental pollution.

2) Abnormal flowmeter

When the flowmeter is abnormal, 
the measured data will be abnormal. For example: 
In sewage networks, 
waste could flow into the pipes and cause blocked flowmeter probe. 
The sewage flow and liquid level data detected by the flowmeter 
will drop sharply in the meantime. 
If the situation is not rectified in a timely manner,
it is likely to cause complete damage to the flowmeter. 
At this time, the flowmeter cannot 
feed back the flow data of the local pipe network.

3) Sewage pipe network blockage

When a large number of garbage and foreign matters enter the sewage pipe network, 
it may cause the blockage of the sewage pipe network. 
At this time, 
the instantaneous flow and liquid level detected by the flowmeter will also decrease. 
If the sewage pipe network can't be dredged in time, 
it will lead to pipe network damage and environmental pollution easily.

4) Network fluctuation and discontinuous transmission

When the sewage pipe network is located in the area with poor signal, 
the data transmission of the sensor corresponding to the pipe network 
will be interrupted due to the unstable signal. 
In the period of time without data, 
the sewer network can’t be monitored effectively, 
there will be some hidden dangers.

5) Full sewage pipe network

When the sewage pipe network is full, 
the instantaneous flow and liquid level are almost zero. 
If the sewage in full pipe state is not treated in time, 
it may lead to explosion of the pipe, 
which will affect the normal operation of sewage pipe network.

\section{Data introduction and Data preprocessing}
\subsection{Data introduction}
According to the data of 136 monitoring points, 
finally 13830 typical data are summarized for model training. 
The data mainly has five dimensions: 
time (day, hour), instantaneous flow ($m^{3}$/h), 
liquid level (m) and flow rate (m/s). 
Among them, instantaneous flow and liquid level 
are the main indicators of diagnostic data in this paper. 
The sampling interval unit of data is 5 minutes. 
When the data is abnormal, abnormal is used to 
identify the data exception. 

Combined with the above five abnormal conditions, 
we classify the data into three abnormal data types:  
sudden change of instantaneous flow and liquid level to zero, 
sudden increase of instantaneous flow and liquid level, 
and sudden decrease of instantaneous flow and liquid level. 
This paper intercepts some data for these three types for display, 
as shown in Tbl. \ref{table1}. \textit{I F} in Tbl. \ref{table1} represents instantaneous flow.

\begin{table}[H]
\newcommand{\tabincell}[2]{\begin{tabular}{@{}#1@{}}#2\end{tabular}}
\centering
\caption{Different exception data types. 
For example, this paper selects the most representative three groups of data (9 / 7, 9 / 12, 4 / 12) in the data set, 
which respectively reflect three different data anomaly types: 
the sudden change of instantaneous flow and liquid level to 0, 
the sudden increase of instantaneous flow and liquid level, 
and the sudden decrease of instantaneous flow and liquid level.}
\begin{tabular}{ccccccc}
\hline
Type     &Day   &Hour    &\tabincell{c}{\textit{I F}\\($m^{3}$/h)}   &\tabincell{c}{liquid level\\($m$)}      &\tabincell{c}{flow rate\\($m/s$)}    &Label \\ \hline
\multirow{6}*{\tabincell{c}{The instantaneous flow\\and liquid level\\suddenly change to zero}}  &9$/$7   &11$:$20  &1$.$288  &0$.$18  &0$.$081  &{/} \\
&9$/$7   &11$:$25  &20$.$929  &0$.$173  &0$.$115  &{/} \\
&9$/$7   &11$:$30  &4$.$326  &0$.$168  &0$.$03  &{/} \\
&9$/$7   &11$:$35  &43$.$969  &0$.$164  &0$.$13  &{/} \\
&9$/$7   &11$:$40  &0  &0  &0  &abnormal \\
&9$/$7   &11$:$45  &0  &0  &0  &abnormal \\
\hline
\multirow{6}*{\tabincell{c}{Sudden increase of\\instantaneous flow\\and liquid level}}  &9$/$12   &7$:$05  &193$.$759  &0$.$291  &0$.$537  &{/} \\
&9$/$12   &7$:$10  &153$.$54  &0$.$281 &0$.$463   &{/} \\
&9$/$12   &7$:$15  &141$.$475  &0$.$265  &0$.$444  &{/} \\
&9$/$12   &7$:$20  &132$.$941  &0$.$255  &0$.$44  &{/} \\
&9$/$12   &7$:$25  &929$.$935  &1$.$237  &2$.$016  &abnormal \\
&9$/$12   &7$:$30  &1478$.$788  &1$.$626  &3$.$286  &abnormal \\
\hline
\multirow{6}*{\tabincell{c}{Sudden decrease of\\instantaneous flow\\and liquid level}}  &4$/$12   &9$:$00  &2103$.$394  &0$.$575  &1$.$079  &{/} \\
&4$/$12   &9$:$05  &2015$.$826  &0$.$58  &1$.$035  &{/} \\
&4$/$12   &9$:$10  &2004$.$826  &0$.$575  &1$.$03  &{/} \\
&4$/$12   &9$:$15  &2007$.$798  &0$.$581  &1$.$026  &{/} \\
&4$/$12   &9$:$20  &755$.$699  &0$.$361  &0$.$738  &abnormal \\
&4$/$12   &9$:$25  &705$.$991  &0$.$357  &0$.$677  &abnormal \\
\hline
\label{table1}
\end{tabular}
\end{table} 

It can be seen in Tbl. \ref{table1} that among the data with 
instantaneous flow and liquid level suddenly changing to zero, 
the instantaneous flow and liquid level suddenly changed around 11:35, 
and after 11:35, the instantaneous flow and liquid level have always been zero. 
Therefore, these data are marked with abnormal label, 
corresponding to the five abnormal types introduced above. 
This phenomenon may be caused by abnormal conditions such as network fluctuations and discontinuous transmission, abnormal flow meters, and full sewage pipe networks. 
As can be seen from the data in Tbl. \ref{table1}, 
the instantaneous flow and liquid level suddenly increased around 7:20. 
So the data after 7:20 is marked with abnormal label, 
corresponding to the five abnormal types described above. 
This phenomenon may be caused by the mixing of external water sources 
and other abnormal conditions. 
When the instantaneous flow and liquid level suddenly decrease, 
it can be seen from the data in the Tbl. \ref{table1} 
that there is a sudden decrease in the instantaneous flow and liquid level 
before and after 9:15. Therefore, the data after 9:15 is marked with abnormal label, 
corresponding to the five abnormal types described above. 
This may be due to blocking the sewage pipe network, 
abnormal flowmeter and other abnormal conditions  

\subsection{Data preprocessing}
\begin{figure}[H]
\centering
\includegraphics[width=0.8\textwidth]{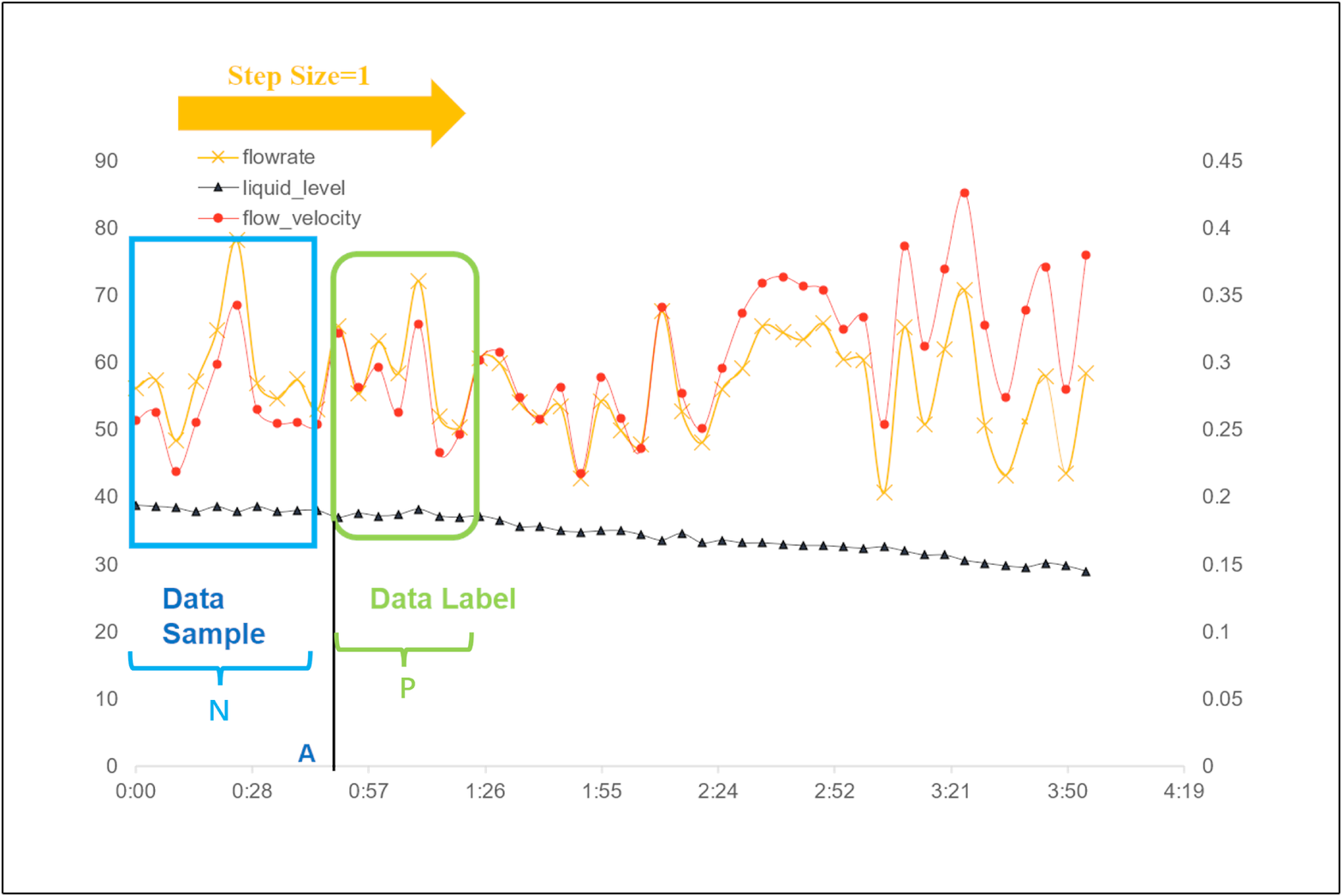}
	\caption{Schematic diagram of data preprocessing}
	\label{data}
\end{figure}

In order to explore the correlation between 
current data and historical data in sewage data, 
and predict the data of future state through historical data, 
this paper preprocesses the data. Among them, 
the research preprocesses the data by sliding window, 
and the operation flow is shown in Fig. \ref{data}.
In Fig. \ref{data}, two sliding windows are set to intercept data. 
Each sliding step of the sliding window is \textbf{1}, 
and the time interval of each step is \textbf{5} minutes. 
The first sliding window intercepts the samples of historical data  
in the \textit{N} time periods before \textit{A} in the figure, 
and the data include instantaneous flow, 
liquid level and flow rate; 
The second sliding window selects the data label starting 
from point a and containing the future data from \textit{A} to \textit{P}. 
If there are abnormal values in \textit{P}, 
it outputs abnormal, otherwise it outputs normal.
\section{Method}
\subsection{Anomaly detection algorithm: A brief review}
Anomaly detection is to detect the abnormal data that does not conform to the behavior in the data, 
and mark its abnormalities. This paper mainly uses three anomaly detection algorithms: 
One Class SVM, Isolation Forest and Local Outlier Factor 
to detect the flow data of sewage pipe network.

\textit{One class SVM.} 
The algorithm is a single classification anomaly detection algorithm given by $Schölkopf$ et al\cite{Sch2001Estimating}. 
The algorithm trains only one sample and detects the unique class in the sample. 
If the class is detected, it returns "yes". If the class is not detected, it returns "no".

The algorithm of one class SVM is as follows. 
If one-dimensional training data $X_{i} \in R^{n} {(i=1,2,3,....,n)}$ is set, there is:
\begin{equation} 
\mathop{\min}_{\omega,i,\xi,\rho} \frac{1}{2}\omega^{T}\omega-\rho+\frac{1}{vn}\sum_{i=1}^n \xi_{i} \tag{1} \label{con:1}
\end{equation}
\begin{equation} 
\left\{
\begin{array}{lr}
\omega^{T}\Phi{(x_{i})}\geq\rho{-}\xi_{i} \\
\xi_{i}\geq{0,}{i=1,2,....,n} 
\end{array}
\right.
\tag{2} \label{con:2}
\end{equation}
The following results are obtained through conversion:
\begin{equation} 
\mathop{\min}_{a}\frac{1}{2}\alpha^{T}Q_{ij}\alpha,\left\{
\begin{array}{lr}
0\leq\alpha_{i}\leq\frac{1}{vn},{i=1,2,....,n} \\
e^{T}\alpha=1
\end{array}
\right.
\tag{3} \label{con:3}
\end{equation}
In formula (3):
\begin{equation} 
Q_{ij}=K(x_{i},x_{j})=\varphi(x_{k})^{T}\varphi(x_{l}) \tag{4} \label{con:4}
\end{equation}
The decision function is:
\begin{equation} 
f(x)=sign{(\sum_{i=1}^nK(x,x_{i}-\rho))} \tag{5} \label{con:5}
\end{equation}

The decision function can calculate the category of data 
and divide the calculation results into two categories. 
The calculation result of 
positive category data is 1 and 
that of negative category data is - 1.\\
The symbol function is:
\begin{equation} 
g(x)=\sum_{i=1}^n \alpha_{i}K(x,x_{i})-\rho \tag{6} \label{con:6}
\end{equation}

This function can judge whether the test data is normal. 
The greater the distance from the test point, 
the greater the deviation of the data from the zero point.

\textit{Isolation Forest.}
The concept of this algorithm is an unsupervised learning proposed by F Liu and other scholars\cite{2008Isolation}, 
which can quickly detect data sets. Compared with other algorithms, 
isolated forest does not need to establish any data model 
and any label description for anomaly detection. 
It only needs to estimate the data in turn to find the outliers.

\begin{figure}[H]
\centering
\includegraphics[width=0.8\textwidth]{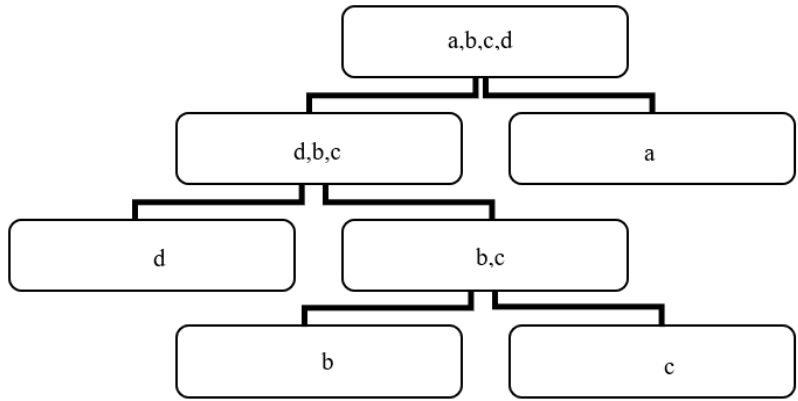}
	\caption{Schematic diagram of isolation forest}
	\label{isolation}
\end{figure}

The isolation forest algorithm puts the data 
on an isolated tree\cite{2012Isolation}, then checks the characteristics of the data in turn, 
and classifies the data with the same characteristics into one class, 
otherwise it is isolated. 
Because there are great differences between abnormal data and normal data in essence, 
abnormal data will be isolated very early. As shown in Fig. \ref{isolation}, 
this figure is an example of isolation forest. 
According to the above description, 
it can be inferred that \textbf{a} was isolated at the earliest 
and is most likely an abnormal point.

\textit{Local Outlier Factor.}
The algorithm is defined by breuing et al.\cite{Breunig2000LOFID}
The algorithm mainly determines whether the data is abnormal 
according to the data density around the data points. 
Firstly, the local reachable density of each data point is calculated, 
and then an outlier factor of each data point is further calculated 
through the local reachable density. 
The outlier factor indicates the outlier degree of a data point. 
When the factor value is larger, it means that the outlier degree is higher, 
which means that the data is abnormal. The smaller the factor value, 
the lower the degree of outlier, and the data is normal.

\begin{figure}[H]
\centering
\includegraphics[width=0.8\textwidth]{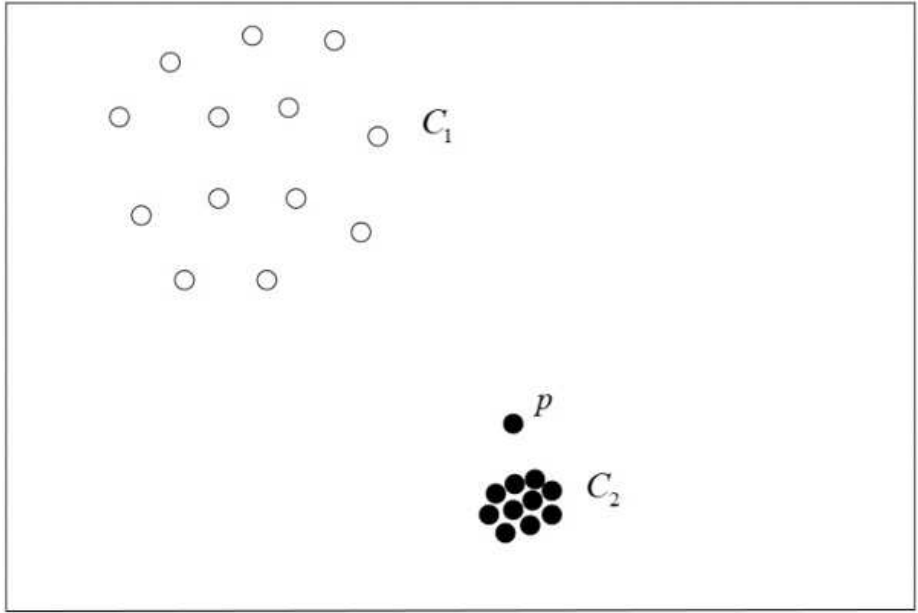}
	\caption{Schematic diagram of local outlier factor}
	\label{Local}
\end{figure}

The idea of local outlier factor algorithm can be shown in Fig. \ref{Local}. 
Set \textit{$C_{1}$} is a low-density area and 
set \textit{$C_{2}$} is a high-density area. 
According to the traditional density based outlier detection algorithm, 
the distance between \textit{p} and the adjacent point in \textit{$C_{2}$} is 
less than the distance between any data point 
in \textit{$C_{1}$} and its adjacent point, 
and \textit{p} will be regarded as a normal point, while locally, 
\textit{p} is actually an isolated point, 
the outlier algorithm can effectively detect outlier in the case of local factor.

\subsection{The bagging based algorithm}
This paper uses the anomaly detection algorithm which is introduced in Chapter 4.1 
to detect the anomaly flow data in Chapter 3.1. 
Among them, the instantaneous flow and liquid level in the data 
are the main indicators of anomaly diagnosis. 
When the abnormal type data shown in Tbl. \ref{table1} appears during the detection, 
the abnormal data is marked with abnormal. 
However, through a large number of experiments, 
it is found that the precision and recall rate of the experimental results 
obtained by the anomaly detection algorithm in Chapter 4.1 are not perfect. 
Therefore, through the integration strategy 
this paper adapt the integrated learning of the anomaly algorithm in Chapter 4.1. 
The idea of ensemble learning is to correct other model results 
through a large number of weak learning model results, 
so as to make the model more consistent 
with the dataset and have a better precision. 

\begin{figure}[H]
\centering
\includegraphics[width=0.9\textwidth]{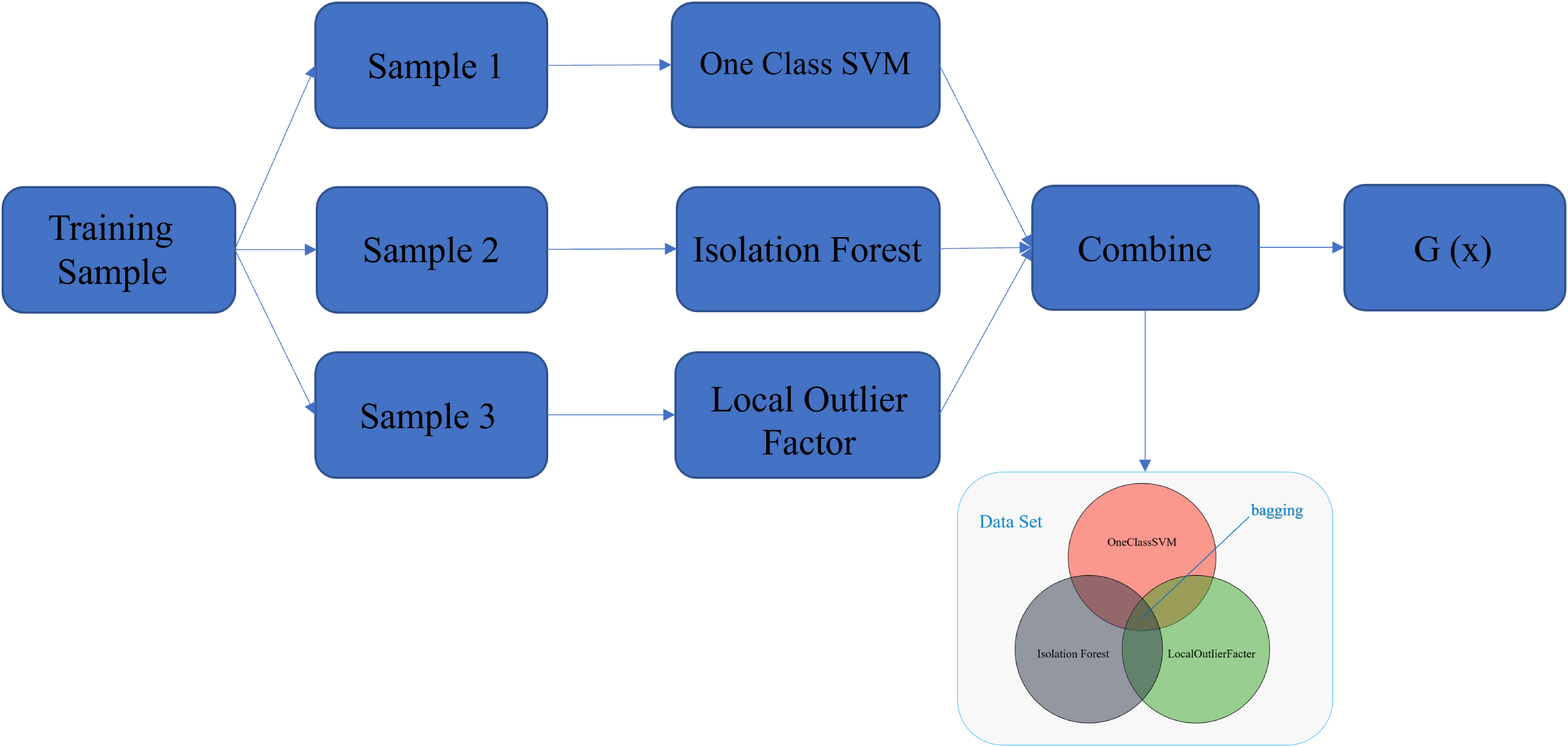}
	\caption{Bagging flow chart}
	\label{bagging}
\end{figure}

The methods of ensemble learning mainly include bagging and boosting. 
The integrated learning method used in this paper is bagging. 
Bagging algorithm \cite{1999An,2003Bagging} uses multiple 
weak learning models to train the randomly collected data sets, 
and then combines the results of multiple weak learning 
to form the final strong learner model. 
The bagging process of this paper is shown in Fig. (\ref{bagging}). 
Training samples are randomly used to train three anomaly detection algorithms: 
one class SVM, isolation forest and local outlier factor. 
Finally, these algorithms are combined by finding the 
combination of the intersection of the three algorithms 
to obtain the integrated algorithm G(x) which is proposed in this paper.
\section{Result}
In this paper, we detect the flow data by the anomaly detection algorithm. 
When the flow data is abnormal, the abnormal data would be marked. 
Therefore, the main purpose of this paper is that explore 
the precision and recall of the algorithm marking abnormal, 
and judge the performance of the algorithm model proposed 
in this paper by marking the precision and recall 
of abnormal and the corresponding F1-score index. 
Among them, F1-score is a measurement index of classification problems, 
referred to as $\emph{F}_{1}$ for short. 
It is the harmonic average of accuracy and recall, 
with the maximum of 1 and the minimum of 0. 
Generally, the larger the index, 
the better the performance of the model. 
The calculation method of $\emph{F}_{1}$ is as follows:
\begin{equation} 
\emph{F}_{1}=2\cdot{\frac{precision\cdot{recall}}{precision+recall}} \tag{7}   \label{con:7}
\end{equation}
If there is a strong correlation between historical data and future data,
the precision and recall rate of the model can be improved. 
In order to get the better correlation, we need to 
find the most appropriate \textit{N} and \textit{P}, 
and this paper explored this problem. As shown in Tbl. \ref{table2}, 
set \textit{N} to 5, 10, 15, 20, 25 and \textit{P} to 7 respectively. 
The comparison shows that when \textit{N}=$5$, 
the integrated anomaly detection algorithm 
has a high precision and recall. 
Therefore, when \textit{N}=$5$, the data of history and future has a good correlation.
\begin{table}[H]
\newcommand{\tabincell}[2]{\begin{tabular}{@{}#1@{}}#2\end{tabular}}
\centering
\caption{Precision and recall of the algorithm in different $N$ cases}
\begin{tabular}{ccccccccc}
\hline
\tabincell{c}{start\\(hour)} & \tabincell{c}{end\\(hour)} & $N$ &$P$ &$abnormal$ &$normal$ &$recall($\%$)$ &$precision($\%$)$ &$\emph{F}_{1}$ \\ \hline
$2$   & $9$  & $5$  & $7$ & $184$  & $546$ & $59.24$ & $99.09$ &$0.742$\\
$2$   & $9$  & $10$ & $7$ & $169$  & $511$ & $59.76$ & $58.06$ &$0.739$\\
$2$   & $9$  & $15$ & $7$ & $154$  & $476$ & $57.79$ & $95.70$ &$0.721$\\
$2$   & $9$  & $20$ & $7$& $139$  & $441$ & $51.08$ & $93.42$ &$0.66$\\ 
$2$   & $9$  & $25$ & $7$ & $126$  & $404$ & $13.49$ & $89.47$ &$0.234$\\
\hline
\label{table2}
\end{tabular}
\end{table} 

When \textit{N}=$5$ is determined, the abnormal data has good correlation before and after. 
Therefore, then looks for an appropriate \textit{P} to 
make the correlation between the abnormal data better. 
As shown in Tbl. \ref{table3}, all of the \textit{N} are $7$, and \textit{P} is set to $5$, $6$, $8$, $9$ and $10$ respectively. 
Through comparison, it can be found that when \textit{N}=$5$ 
and \textit{P}=$5$, 
the precision and recall of the integrated anomaly detection algorithm are better. 
Therefore, when \textit{N}=$5$ and \textit{P}=$5$, the correlation of 
historical data and future data is the best. 
\begin{table}[H]
\newcommand{\tabincell}[2]{\begin{tabular}{@{}#1@{}}#2\end{tabular}}
\centering
\caption{Precision and recall of the algorithm in different $P$ cases}
\begin{tabular}{ccccccccc}
\hline
\tabincell{c}{start\\(hour)} & \tabincell{c}{end\\(hour)} & $N$ &$P$ &$abnormal$ &$normal$ &$recall($\%$)$ &$precision($\%$)$ &$\emph{F}_{1}$\\ \hline
$2$   & $9$  & $5$ & $5$  & $173$  & $577$ & $63.58$ & $98.21$ &$0.772$\\
$2$   & $9$  & $5$ & $6$  & $179$  & $561$ & $61.45$ & $98.21$ &$0.756$\\
$2$   & $9$  & $5$ & $8$  & $189$  & $531$& $37.04$ & $100$ &$0.541$\\
$2$  & $9$  & $5$ & $9 $ & $194$  & $516$ & $35.57$ & $100$ &$0.525$\\ 
$2$   & $9$  & $5$ & $10$ & $198$  & $502$ & $34.34$ &$ 100$ &$0.511$\\
\hline
\label{table3}
\end{tabular}
\end{table} 
When \textit{N}=$5$ and \textit{P}=$5$, 
the correlation of abnormal data is the best. 
Therefore, in the case of \textit{N}=$5$ and \textit{P}=$5$, 
this paper trains the three anomaly detection algorithms, 
adjusts the parameters, and then integrates 
the three anomaly detection algorithms through using the integration strategy. 
Finally, the integrated experimental results are obtained.

\begin{figure}[H]
\centering
\includegraphics[width=0.8\textwidth]{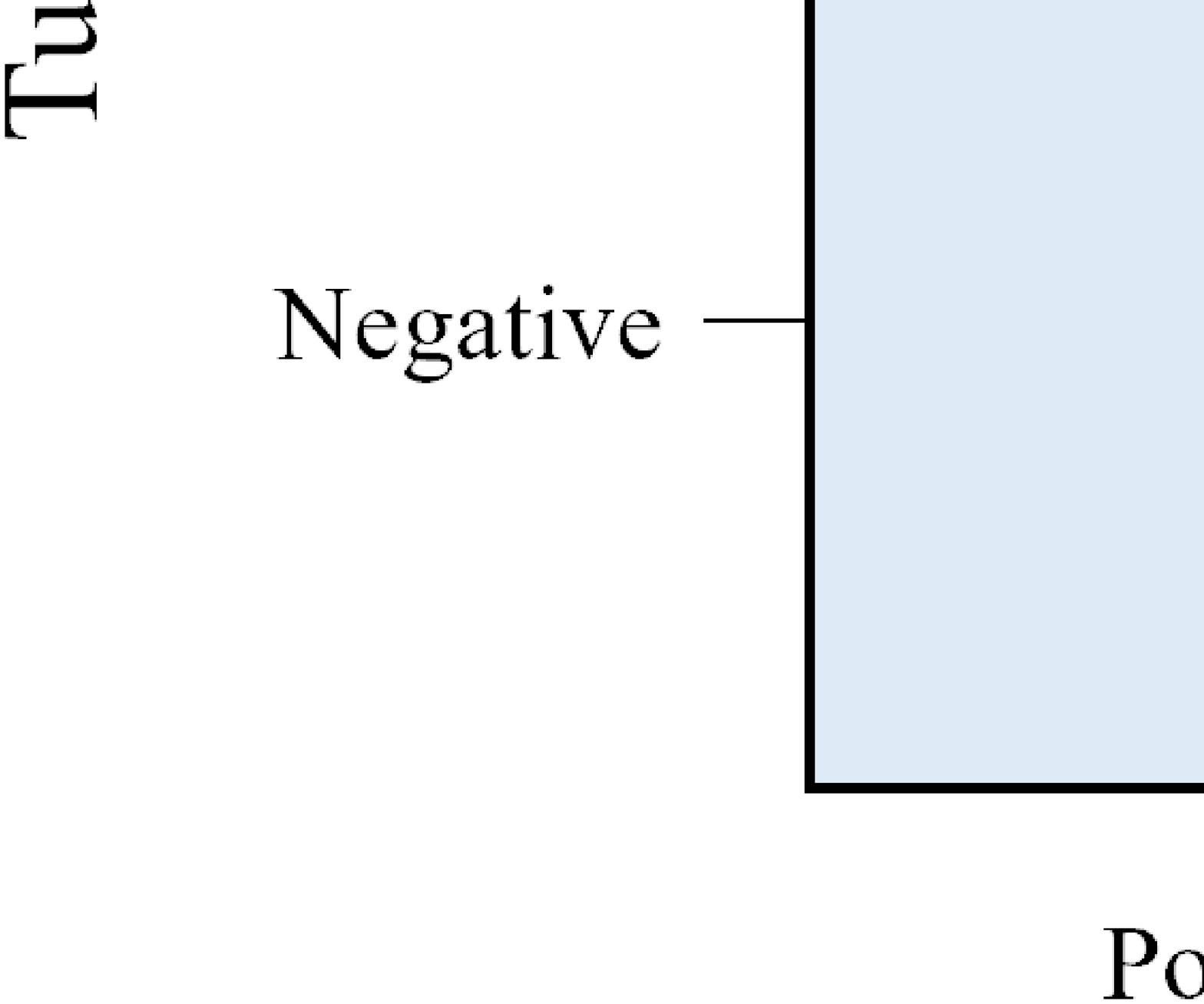}
	\caption{Confusion matrix of three anomaly detection algorithms}
	\label{one}
\end{figure}

\begin{figure}[H]
\centering
\includegraphics[width=0.8\textwidth]{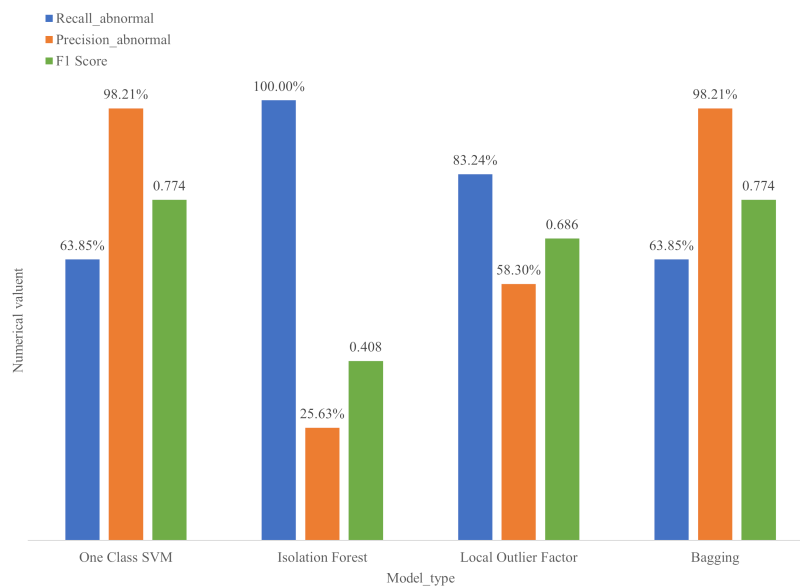}
	\caption{Comparison of results of three anomaly detection algorithms and integrated algorithms}
	\label{one1}
\end{figure}

In Fig. \ref{one}, a to c are the confusion matrix obtained 
after training of One Class SVM, Isolation Forest and Local Outlier Factor. 
The precision and recall rates corresponding to the three anomaly detection algorithms 
and the integrated algorithm are shown in Tbl. \ref{table4}, 
and the comparison diagram of the results is shown in Fig. \ref{one1}. 
The experimental results show that the recall rate of the integrated 
algorithm is 63.85$\%$ and the precision rate is 98.21$\%$. 
It's an effective anomaly detection algorithm 
for early warning of abnormal sewage flow data.
\begin{table}[H]
\centering
\caption{Recall and precision of three anomaly detection algorithms and integrated algorithms}
\begin{tabular}{cccc}
\hline
$model$ & $recall($\%$)$ & $precision($\%$)$ &$\emph{F}_{1}$\\ \hline
$One Class SVM$  &$63.85$   & $98.21$ &$0.774$\\
$Isolation Forest$   & $100$  & $25.63$ &$0.408$\\
$Local Outlier Factor$ & $83.24$  & $58.30$ &$0.686$\\
$Bagging$ & $63.85$  & $98.21$ &$0.774$\\ 
\hline
\label{table4}
\end{tabular}
\end{table} 

\begin{figure}[H]
\centering
\includegraphics[width=0.8\textwidth]{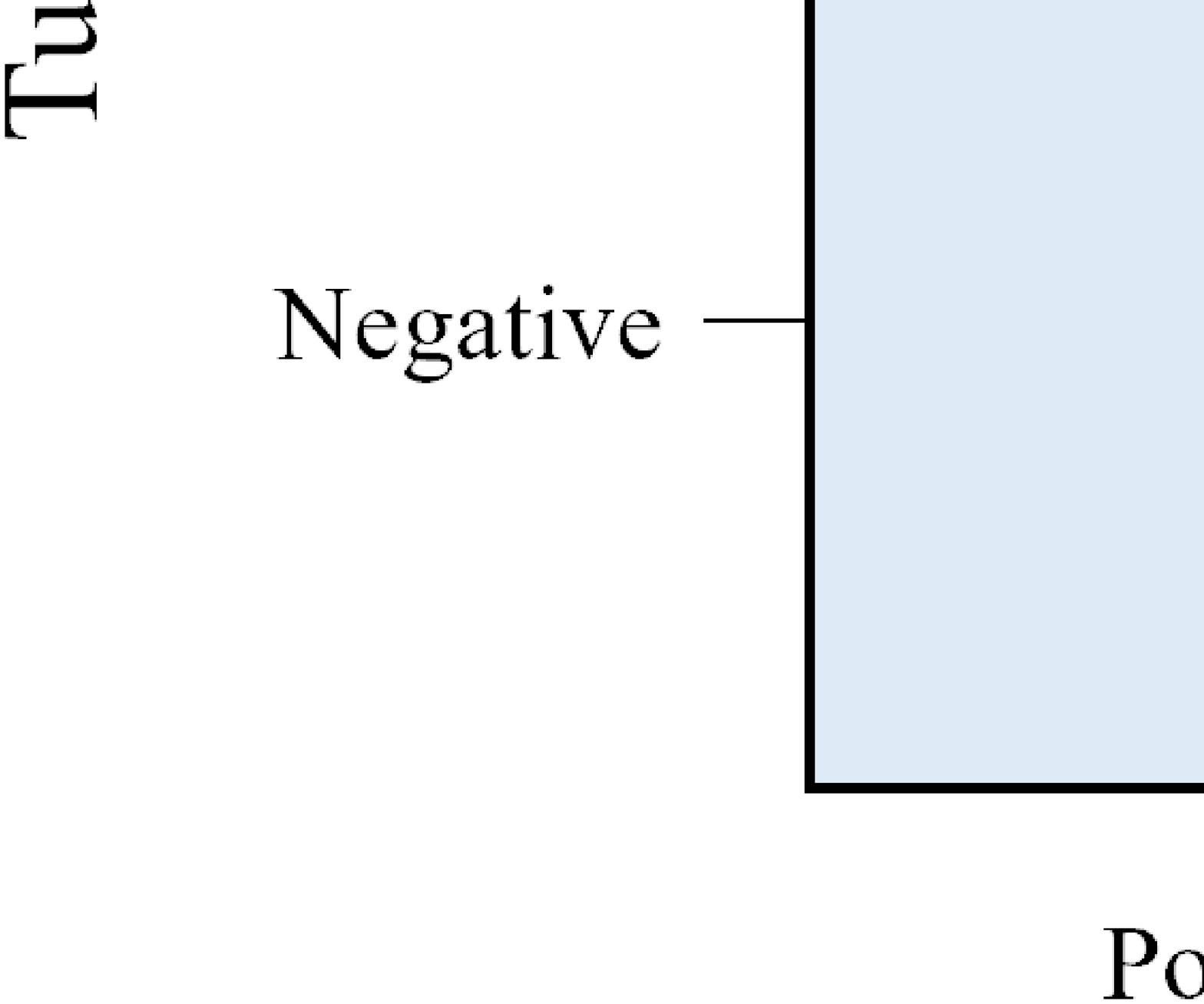}
	\caption{Confusion matrix of three anomaly detection algorithms}
	\label{two}
\end{figure}

\begin{figure}[H]
\centering
\includegraphics[width=0.8\textwidth]{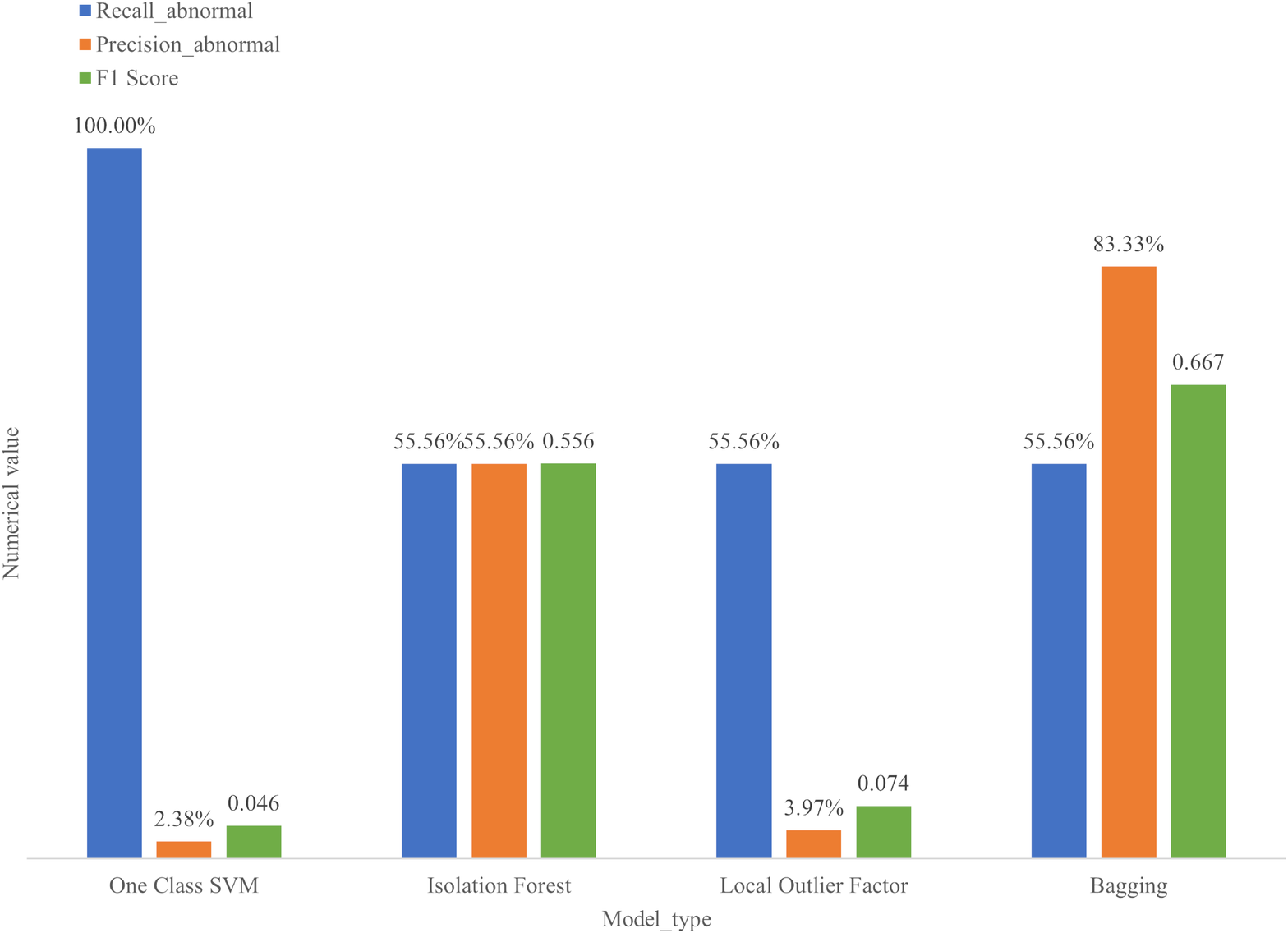}
	\caption{Confusion matrix of three anomaly detection algorithms}
	\label{two2}
\end{figure}

When the predicted results of the three anomaly detection algorithms are not good, 
the integrated algorithm proposed in this paper 
can significantly improve the precision while retaining a high recall rate. 
As shown in Tbl. \ref{six}, it can be seen that although 
the integrated algorithm loses little recall rate, 
compared with One Class SVM, 
the precision of the algorithm is significantly improved 
and the false alarm rate of abnormal sewage flow is greatly reduced. 
The confusion matrix and result comparison diagram of three anomaly detection algorithms 
and integrated algorithms are shown in Fig. \ref{two} and Fig. \ref{two2} respectively.

\begin{table}[H]
\centering
\caption{Recall and precision of three anomaly detection algorithms and integrated algorithms}
\begin{tabular}{cccc}
\hline
$model$ & $recall($\%$)$ & $precision($\%$)$ &$\emph{F}_{1}$\\ \hline
$One Class SVM$  &$100$   & $2.38$ &$0.046$\\
$Isolation Forest$   & $55.56$  & $55.56$ &$0.556$\\
$Local Outlier Factor$ & $55.56$  &$ 3.97$ &$0.074$\\
$Bagging$ & $55.56$  & $83.33$ &$0.667$\\ 
\hline
\label{six}
\end{tabular}
\end{table} 

Through the above experimental results, 
it can be seen that the unsupervised anomaly detection integration algorithm proposed 
in this paper is an effective and good performance method, 
which can effectively warn the abnormal sewage flow data.

\section{Conclusions and Outlooks}
This paper proposed a abnormal-detection algorithm. 
Based on the original hardware facilities, 
the anomaly detection algorithm is combined with the data 
obtained by the hardware facilities to form a feasible, 
low-cost and effective sewage pipe network monitoring system. 
The data of this project comes from the 
intelligent monitoring project of The Pollution Control of Erhai Lake. 
The anomaly detection algorithm which is proposed in this paper 
is trained by using these data, 
and we can get a better result that the precision is 98.21$\%$, recall rate is 63.85$\%$, and F1-score is $0.774$. 
It’s a more effective anomaly detection algorithm 
to detect and warn when the flow data of sewage pipe network is abnormal. 
This paper has completed the design of abnormal early warning system 
of sewage pipe network from the aspects of theory and practice, 
and achieved a good results in practical application and simulation scenarios. 
However, there are some other defect in this research of 
abnormal early warning of sewage pipe network, 
it can be further carried out in future research:

1) Implementation of abnormal classification and early warning algorithm 
of sewage pipe network. 
The abnormal early warning system of sewage pipe network has been finished, 
but there is no good result for the classification of abnormal points. 
In the subsequent research, we can collect a large number of abnormal data 
of sewage pipe network to train the classification algorithm of abnormal sewage flow, 
and finish the task of abnormal classification 
and early warning algorithm of sewage pipe network.

2) Improve the precision of sewage anomaly early warning algorithm. 
Although this project has improved the precision of anomaly early warning to 98.21$\%$, 
and recall rate also could be better. 
Through the lucubration and updates of the model, 
we can improve the recall rate and precision of sewage anomaly early warning algorithm.

3) This project mainly focuses on the early warning of abnormal sewage flow in non full pipe, 
As for full pipe,this project don’t involve too much. 
Therefore, the follow-up study can aim at abnormal early warning 
for the full pipe of sewage flow.

\section{Acknowledgments}
This project get the strongly support 
from the intelligen monitoring project of The Pollution Control of Erhai Lake. 
Chi-Chun Zhou acknowledges the supported by National Natural Science Funds of China (Grant No. 62106033), 
Yunnan Youth Basic Research Projects (202001AU070020 and 202001AU070022), 
and Doctoral Programs of Dali University (KYBS201910).
En-Ming Zhao acknowledges the supported by National Natural Science Funds of China (Grant No. 62065001).


\end{document}